\RequirePackage{fix-cm}
\documentclass[smallextended]{svjour3}       \smartqed  

\usepackage{tikz}
\usepackage{pgfplots}

\usepackage{graphicx}

\usepackage{amsmath}
\usepackage{amssymb}
\usepackage{booktabs} 
\usepackage{todonotes}
\usepackage{nicefrac}
\usepackage{subcaption}

\usepackage{siunitx}
\usepackage{multirow, hhline}

\usepackage{url}

\usepackage{tikz}
\usetikzlibrary{math, positioning, patterns, matrix, decorations.pathreplacing, calc}
\usepackage{pgfplots}
\usepackage{pgfplotstable}
\usepgfplotslibrary{fillbetween}
\pgfplotsset{
compat=1.6,
}

\usepackage{natbib}

\usepackage{enumitem}
\usepackage{array, multirow}

\definecolor{darkgreen}{rgb}{0.0,0.5,0.0}
\definecolor{amber}{rgb}{0.8, 0.6, 0.0}

\DeclareMathOperator*{\argmin}{arg\,min}
\DeclareMathOperator{\diag}{diag}
\DeclareMathOperator{\rank}{rank}

\def\makeheadbox{}

\begin{document}

\title{Pseudoinverse Graph Convolutional Networks}
\subtitle{Fast Filters Tailored for Large Eigengaps of Dense Graphs and Hypergraphs}

\author{Dominik Alfke \and Martin Stoll}

\institute{
	Dominik Alfke (corresponding author) \and Martin Stoll \at 
	Chemnitz University of Technologies, Chemnitz, Germany \\
	\email{\{alfke, stoll\}@math.tu-chemnitz.de}\\
}

\date{January 26, 2021}

\maketitle

\begin{abstract}
Graph Convolutional Networks (GCNs) have proven to be successful tools for semi-supervised classification on graph-based datasets. We propose a new GCN variant whose three-part filter space is targeted at dense graphs. Our examples include graphs generated from 3D point clouds with an increased focus on non-local information, as well as hypergraphs based on categorical data of real-world problems. These graphs differ from the common sparse benchmark graphs in terms of the spectral properties of their graph Laplacian. Most notably we observe large eigengaps, which are unfavorable for popular existing GCN architectures. Our method overcomes these issues by utilizing the pseudoinverse of the Laplacian. Another key ingredient is a low-rank approximation of the convolutional matrix, ensuring computational efficiency and increasing accuracy at the same time. We outline how the necessary eigeninformation can be computed efficiently in each applications and discuss the appropriate choice of the only metaparameter, the approximation rank.
We finally showcase our method's performance regarding runtime and accuracy in various experiments with real-world datasets.
\end{abstract}

\vspace{1em}

%
%
%
%
%
%
%

\begin{acknowledgements}
	D. Alfke gratefully acknowledges partial funding by the S\"achsische Aufbaubank -- F\"orderbank -- (SAB) 100378180.
\end{acknowledgements}

\section{Introduction}
\label{sec:intro}

One of the central tasks in data science applications is extracting information from relational data encoded in graphs. 
The umbrella term Graph Neural Networks (GNNs) comprises neural models that seek to combine the theoretical understanding of structured data with the flexibility of machine learning. An outstandingly successful class of GNNs relies on spectral convolution of features along the graph edges \citep{bruna14,defferard16}. These Graph Convolutional Networks (GCNs) have become particularly popular through their success in semi-supervised node classification tasks \citep{kipf17}, where methods aim to benefit from both a small set of training data and clustering information extracted from a large amount of unlabeled data.

In this work, we consider \emph{dense} graphs where any node is connected to most other nodes. This structure occurs in several applications where each node of the graph describes a real-world entity but the edges are constructed artificially based on specific node features. While it is possible to represent such examples using sparse networks this requires additional model assumptions that can be easily avoided using our approach.  We here discuss in detail two relevant examples of such constructed, dense graphs. First, a Gaussian kernel function can be used to generate edge weights for fully connected graphs based on spatial node features, e.g., for three-dimensional point clouds as created by LiDAR scans \citep{nguyen13pointclouds}. A localization parameter determines how fast the weights decay with the spatial distance, which can be understood to control the density of the graph. Up to now, only approaches with sparse $k$-nearest neighbor graphs have been proposed, but these are always associated with the loss of non-local information encoded within the large number of edges with lesser weights. We will show that increasing the density of the graph improves the prediction performance. This phenomenon is also well-known in other related fields like computer vision \citep{coll05nonlocal, tao18nonlocalnn} and image processing \citep{gilboa08nonlocal}.
For a second example, \emph{hypergraphs} provide a natural extension of graph learning that can be used easily for categorical data \citep{bretto13}. For the purpose of many successful methods, these hypergraphs are equivalent to specific dense graphs with exploitable structure.
Both these types of constructed graphs consist of edges that do not directly represent real-world connections but still encode valuable information about the dataset.

The density of constructed graphs has far-reaching consequences on the performance of GNNs.
Graph learning has traditionally been targeted at purely data-inherent graphs where not only the nodes describe real-world entities, but also every single edge represents a real-world connection.
Since the number of edges per node is often limited by real-world factors independent of the network size, the average node degree in these graphs is asymptotically constant, the total number of edges grows linearly with the network size, and the adjacency matrix is sparse. 
When moving to dense graphs, the increased computational cost and storage requirements might be expected to be a decisive issue. That is however often not the case, as the special structure of the constructed adjacency matrices can be exploited to speed up the relevant algorithms.
The true problem is posed by
the intrinsically different spectral properties of 
the dense graph Laplacian.
It is well known in graph theory that the smallest eigenvalue $\lambda_0$ is always zero and the second smallest eigenvalue $\lambda_1$ gives a measure of how close the graph is to not being connected \citep{bauer09}. We call $\lambda_1$ the graph's \emph{eigengap} as the difference between the first and second eigenvalue.
The eigenvectors corresponding to the first nonzero eigenvalues are known to contain clustering information. 
Most common GNNs reinforce this clustering information through their feature maps. In \emph{spectral} approaches, this is achieved explicitly via convolution with a spectral filter designed specifically for that purpose. 
However, denser graphs empirically have much larger eigengaps and almost all eigenvalues are clustered close to 1 \citep{bauer09}.
For that reason, the informative eigenvectors are considerably harder to extract by existing filters as well as spatial feature maps. Common GNN architectures hence tend to underperform on dense graphs while our method embraces density and its spectral properties.

In this setting, the present work offers the following main contributions:
\begin{itemize}
	\item We give motivation for spectral filters that combine a zero-impulse part and an inverse part to overcome the issues related with large eigengaps.
	In order to make our approach computationally efficient, we add a high-pass part and employ low-rank approximations to the pseudoinverse by computing a small number of eigenvalues of the graph Laplacian.
	
	\item We propose a Graph Convolutional Network architecture with a three-dimensional filter function space that represents learning our filter parameters in training. We discuss computational aspects such as asymptotic cost and parameter influence.
	
	\item We consider two examples for applications where beneficial dense graphs can be constructed. We discuss how the intrinsic structure of these graphs can be exploited to speed up our setup.
	
	\item We showcase the performance of our method in various experiments, comparing it to recent popular GNNs.
	
\end{itemize}

\subsection{Related work}
\label{sec:intro:related}

\paragraph{Graph Neural Networks.}
Neural networks have been used for learning on graphs for many years and the vast variety of recent methods and extensions can best be 
studied from the
dedicated review articles \citep{bronstein17, wu19, zhang19}. As for methods with a particular connection to our work, we would like to single out Graph Diffusion Convolution \citep[GDC;][]{klicpera19} and ARMA filters \citep{bianchi19}. These methods also perform convolution with non-linear spectral filters, in GDC even non-locally. However, their filters are not specifically designed towards boosting clustering information
in dense graphs.

\paragraph{GNN techniques for dense graphs.}
To the best of our knowledge, the challenges posed by dense graphs have rarely been addressed in the context of GNNs.
In GDC \citep{klicpera19}, however, a graph is transformed into a dense graph whose adjacency matrix incorporates diffusion information. This dense graph is then sparsified by a $k$-Nearest Neighbor approach, which is argued to both address runtime issues and improve prediction accuracy empirically. This sparsification process can equally be applied to general dense graphs, though it has to be noted that non-local information will be lost. 
Other techniques for increased efficiency like, e.g., node sampling \citep{hamilton17,chen18} are explicitly targeted at large, sparse graphs.

\paragraph{Semi-supervised classification on hypergraphs.}
Hypergraph generalizations of the graph Laplacian operator have been introduced in competing ways. A definition based on the clique expansion graph \citep{zhou06} has been used for various approaches based on energy minimization \citep{hein13,bosch18} as well as Hypergraph Neural Networks \citep{feng19}, a natural generalization of GCNs. A similar work targets hypergraphs constructed from graphs \citep{bai19}. \citet{yadati19} quote density issues as motivation to avoid the clique expansion graph and instead propose HyperGCN using the Laplacian definition as a nonlinear diffusion operator introduced by \citet{chan18}. None of these works address the Laplacian structure resulting from categorical data.

\paragraph{Inverse Laplacians in Data Science.}
\citet{herbster05online} use the  pseudoinverse of the graph Laplacian for online learning on graphs. Otherwise, mainly inverses of shifted Laplacians can be found in the literature.
On multi-layer graphs, higher negative powers of shifted Laplacians can be used to form a Power Mean Laplacian \citep{mercado18powermean}.
In the context of GNNs, the approximated inverse of a shifted Laplacian is the heart of the diffusion operator for Personalized Page-Rank in GDC \citep{klicpera19}. In a broader sense, it is also an example of rational filtering, which is the basis of ARMA networks \citep{bianchi19}.

\subsection{Problem setting and terminology}
\label{sec:intro:setting}

Throughout this paper, we assume that we are given an undirected weighted graph whose $n$ nodes form the samples of the dataset, and each sample is associated with a $d$-dimensional feature vector. We assume that the graph is connected and non-bipartite, 
which is trivially fulfilled in the setting of dense graphs. On this data we aim to perform semi-supervised node classification. The goal is to assign one of $m$ classes to each sample in such a way that nodes with a strong connection in the graph are likely to belong to the same class. For a small subset of samples called the training set, the true class is known a priori.

The graph can be described mathematically by its weighted adjacency matrix $A \in \mathbb{R}^{n \times n}$, where $A_{ij}$ holds the weight of the edge between nodes $i$ and $j$. This is set to 0 if the nodes are not connected. The degree matrix $D \in \mathbb{R}^{n \times n}$ is defined as the diagonal matrix holding the node degrees $D_{ii} = \sum_{j=1}^n A_{ij}$.
Together they can be used to create the symmetrically normalized graph Laplacian matrix
\[
	\mathcal{L}_{\mathrm{sym}} = I - D^{-1/2} A D^{-1/2},
\]
where $I$ is the identity matrix. $\mathcal{L}_{\mathrm{sym}}$ is known to encode many useful clustering properties of the graph \citep{luxburg07}. 

Spectral graph theory now utilizes the eigenvalue decomposition of the graph Laplacian. Let $(\lambda_i, u_i)$ be the eigenpairs of $\mathcal{L}_{\mathrm{sym}}$ for $i=0,\ldots,n-1$ such that $\mathcal{L}_{\mathrm{sym}} u_i = \lambda_i u_i$ and $\|u_i\|_2=1$. 
Since $\mathcal{L}_{\mathrm{sym}}$ is symmetric, we have 
\[
	\mathcal{L}_{\mathrm{sym}} = U \Lambda U^T 
	\quad \text{with} \quad 
	\Lambda = \begin{bmatrix} \lambda_0 & & \\ & \ddots & \\ & & \lambda_{n-1} \end{bmatrix} 
	\quad \text{and} \quad 
	U = \begin{bmatrix} \vert & & \vert \\ u_0 & \cdots & u_{n-1} \\ \vert & & \vert \end{bmatrix}.
\]
It is known that the smallest eigenvalue is always 0 with multiplicity one since there is only one connected component, and the largest eigenvalue is less than 2 since the graph is non-bipartite \citep{luxburg07}. We assume the eigenvalues to be sorted increasingly, $0 = \lambda_0 < \lambda_1 \leq \ldots \leq \lambda_{n-1} < 2$.
The corresponding eigenvector to $\lambda_0=0$ is $u_0$ with entries equal to the square roots of the node degrees,
\begin{equation}
\label{eq:firsteigenvector}
u_0 = \left[ D_{11}^{1/2},\ \ldots,\ D_{nn}^{1/2} \right]^T.
\end{equation}
In case of multiple connected components, all eigenvectors for $\lambda_0$ can be computed in a similar fashion.

\paragraph{Convolution on graphs.}
The generalization of Fourier transforms and convolution from regular grids to irregular graphs is a central result of spectral graph theory \citep{shuman13}. Let $x \in \mathbb{R}^n$ be a spatial graph signal, i.e., $x_i$ describes the magnitude of some quantity on node $i$. Then the function $\hat{x}(\lambda_j) = u_j^T x$, whose domain is the spectrum of the graph Laplacian, is the result of the graph Fourier transform. Its inverse is $x = \sum_{j=0}^{n-1} \hat{x}(\lambda_j) u_j$. In the spectral space, convolution with another spectral element $\varphi$ is simply point-wise multiplication, $\hat{y}(\lambda_j) = \varphi(\lambda_j) \hat{x}(\lambda_j)$. Note that $\varphi$ must be a real-valued function defined on the eigenvalues $\lambda_j$, but for simplicity, it is commonly defined on the whole interval $[0,2]$. Put together, the spatial representation of the convolution of a spatial signal $x$ with a spectral filter $\varphi$ turns out to be $y = U \varphi(\Lambda) U^T x$, where $\varphi(\Lambda)$ denotes the diagonal matrix holding the function values of $\varphi$ in the diagonal elements of $\Lambda$. The operator $\mathcal{K} = U \varphi(\Lambda) U^T$ is sometimes called the convolutional matrix associated with $\varphi$.

\paragraph{Absence of loops.}
Some popular GNN methods preprocess the graph by adding loops (edges with the same start and end node) with a uniform weight. For GCN, this has been interpreted as a \emph{re-normalization trick} \citep{kipf17}. Besides that interpretation, the self loops empirically lead to slightly reduced eigengaps and hence better performance of the traditional methods that benefit from small eigengaps. This is because the normalized weight of non-loop edges is decreased, making the graph slightly sparser from a spectral point of view. 
Since this behaviour is not required in our method, we generally do not use loops.

\section{Proposed architecture}
\label{sec:framework}

\subsection{Pseudoinverse spectral filter functions}
\label{sec:framework:pinv}

Many popular machine learning methods involve repeated multiplication of a feature matrix with some kind of adjacency matrix. 
The most common form of GCN uses the normalized adjacency matrix, $\hat{A} = D^{-1/2} A D^{-1/2}$, as the convolutional matrix with the spectral filter $\varphi(\lambda) = 1-\lambda$ \citep{kipf17}.
If all non-zero eigenvalues are close to 1, then $\varphi(\lambda)$ is close to zero for all eigenvalues except the first one. Hence $\hat{A} x = U \varphi(\Lambda) U^T x$ will be dominated by the first eigenvector $u_0$ for most $x$, i.e.,
$\hat{A} x$ will almost be a multiple of $u_0$ and close to orthogonal to all other $u_i$ ($i>0$).
This can be seen via the inner product
\[
	u_i^T \hat{A} x = \varphi(\lambda_i) u_i^T x \ \begin{cases} = u_0^T x & \text{if } i = 0, \\ \approx 0 & \text{else.} \end{cases}
\]
Since all entries of $u_0$ have the same sign, this makes it hard for the network output to contain meaningful clustering information.

In order to overcome the issues of multiplying with the adjacency matrix, we would like to introduce the \emph{pseudoinverse} of the graph Laplacian, denoted by $\mathcal{L}_{\mathrm{sym}}^\dagger$. This has been used, e.g., for online learning \citep{herbster05online}.
At the same time, $\mathcal{L}_{\mathrm{sym}}^\dagger$ is also the convolutional matrix of the spectral filter function \citep[cf.][]{golubvanloan}
\[
	\varphi^\dagger(\lambda) = \begin{cases} 0 & \text{if } \lambda = 0, \\ \frac{1}{\lambda} & \text{if } \lambda > 0. \end{cases}
\]
This function is decreasing on the nonzero eigenvalues, so low-frequency signals are reinforced and high-frequency signals are damped.
Because of $\varphi(0) = 0$, the first eigenvector $u_0$ is completely removed from the output of the convolution $\mathcal{L}_{\mathrm{sym}}^\dagger x$, i.e., that output will always be orthogonal to $u_0$. The problematic behaviour described above is hence completely avoided by the pseudoinverse filter.
Note that the decay on the positive eigenvalues can also be achieved by different filters that might warrant investigation in the future.

However, this is a double-edged sword, since often some limited presence of $u_0$ may be beneficial. Hence we propose custom filters that allow for the combination of the pseudoinverse approach with added $u_0$. This can be achieved by filter functions of the form
\begin{equation}
\label{eq:filter2d}
	\varphi_{\alpha,\beta}(\lambda) = \begin{cases} \alpha & \text{if } \lambda = 0, \\ \frac{\lambda_1 \beta}{\lambda} & \text{if } \lambda > 0, \end{cases}
\end{equation}
where the parameters $\alpha,\beta$ can either be assigned manually or learned. The eigengap $\lambda_1$ appears in the pseudoinverse part as a normalization factor. The corresponding convolutional matrix is
\[
	\mathcal{K}_{\alpha,\beta} = U \varphi_{\alpha,\beta}(\Lambda) U^T = \alpha u_0 u_0^T + \lambda_1 \beta \mathcal{L}_{\mathrm{sym}}^\dagger.
\]

\subsection{Low-rank approach}
\label{sec:framework:lowrank}

In most scenarios we cannot compute the full eigenvalue decomposition of $\mathcal{L}_{\mathrm{sym}}$. Instead, we compute only the $r+1$ smallest eigenvalues and replace the term $\beta/\lambda$ in \eqref{eq:filter2d} with a third parameter $\gamma$ for all $\lambda > \lambda_r$. 
This means switching to a high-pass filter for higher frequencies, leading to the filter function
\begin{equation}
\label{eq:filter3d}
	\varphi_{\alpha,\beta,\gamma,r}(\lambda) = \begin{cases} 
	\alpha & \text{if } \lambda = 0, \\
	\frac{\lambda_1 \beta}{\lambda} & \text{if } 0 < \lambda \leq \lambda_r, \\
	\lambda_1 \gamma & \text{if } \lambda > \lambda_r \end{cases}
\end{equation}
and the corresponding convolutional matrix
\[
	\mathcal{K}_{\alpha,\beta,\gamma,r} = 
(\alpha - \lambda_1 \gamma) u_0 u_0^T + \lambda_1 U_r (\beta \Lambda_r^{-1} - \gamma I) U_r^T + \lambda_1 \gamma I,
\]
where 
\[
	\Lambda_r = \begin{bmatrix} \lambda_1 && \\ &\ddots& \\ && \lambda_r \end{bmatrix} \in \mathbb{R}^{r \times r}
	\quad \text{and} \quad
	U_r  = \begin{bmatrix} \vert && \vert \\ u_1 & \cdots & u_r \\ \vert && \vert \end{bmatrix} \in \mathbb{R}^{n \times r}
\]
denote the matrices holding the second through $(r+1)$-st eigenvalues and corresponding eigenvectors in an ascending order. Note that $U_r \Lambda_r^{-1} U_r^T$ is the best rank-$r$ approximation to $\mathcal{L}_{\mathrm{sym}}^\dagger$, i.e.,
\[
	U_r \Lambda_r^{-1} U_r = \argmin_{A \in \mathbb{R}^{n\times n}, \ \rank(A) \leq r} \|\mathcal{L}_{\mathrm{sym}}^\dagger - A\|_2,
\]
which can be proven following the argumentation by, e.g., \citet[Section 7.4.2]{hornjohnson85}.

On the one hand, this low-rank approach is necessary to avoid computing the full eigenvalue decomposition. On the other hand, it
also has spectral benefits. 
In spectral graph theory for clustering applications, the smallest, but nonzero eigenvalues $\lambda_1,\lambda_2,\ldots$ are called the \emph{informative} eigenvalues. Their quantity depends on the graph. The corresponding eigenvectors contain clustering information in the sense that if two nodes have a strong connection in the graph, the corresponding entries in the eigenvector are likely to be similar, especially in terms of their sign.
If $r$ is chosen 
appropriately,
this low-rank filter can be argued to perform pseudoinverse convolution on the informative part of the spectrum, while the non-informative eigenvectors -- especially noise -- are damped uniformly.
Hence, the low-rank approximation may very well have positive effects on the accuracy.
Since there is no hard boundary between informative and non-informative eigenvalues, 
there is a wide range of ranks that can yield these benefits.

\subsection{Pseudoinverse filters in Graph Convolutional Networks}
\label{sec:framework:featuremap}

The general idea of convolutional feature maps in GNNs is that each column $y_j$ of an output matrix $Y \in \mathbb{R}^{n \times N_1}$ is obtained by convolving each column $x_i$ of an input matrix $X \in \mathbb{R}^{n \times N_0}$ with its own learned filter $\varphi_{ij}$ and then summing up the convolution results. The learned filters are restricted to a given function space, which is arguably 
the decisive property of each convolutional GNN variant.
Let that filter space be $K$-dimensional and spanned by the basis functions $\varphi^{(1)},\ldots,\varphi^{(K)}$. Then the coefficients of $\varphi_{ij}$ in the given basis are the trained layer parameters, denoted by $W_{ij}^{(1)}, \ldots, W_{ij}^{(K)}$. The individual filter functions are given as
\begin{equation}
\label{eq:learnedfilter}
	\varphi_{ij} = \sum_{k=1}^K W_{ij}^{(k)} \varphi^{(k)}.
\end{equation}
By organising the parameters in matrices $W^{(k)} \in \mathbb{R}^{N_0 \times N_1}$, this leads to the formula \citep{bruna14, defferard16}
\begin{equation*}
	y_j = \sum_{i=1}^{N_0} U \varphi_{ij}(\Lambda) U^T x_i = \sum_{i=1}^{N_0} \sum_{k=1}^K W_{ij}^{(k)} \underbrace{U \varphi^{(k)}(\Lambda) U^T}_{= \mathcal{K}^{(k)}} x_i,
\end{equation*}
and by putting together the full input feature matrix $X$ with columns $x_i$ and the output feature matrix $Y$ with columns $y_j$, we obtain the feature map
\begin{equation}
\label{eq:pgcn:featuremap}
Y = \sum_{k=1}^K \mathcal{K}^{(k)} X W^{(k)}.
\end{equation}
We now aim to use a small filter space that contains the proposed pseudoinverse filters.
One possibility is to choose the parameters in \eqref{eq:filter3d} manually and set the resulting $\varphi_{\alpha,\beta,\gamma,r}$ as the only basis function for a one-dimensional filter space. That requires a-priori identification of the parameter impact on desirable behaviour, which differs from dataset to dataset. For that reason, this approach is infeasible in practice.

Instead,
we only fix $r$ in \eqref{eq:filter3d} and
have the other parameters learned in training,
which means using the filter basis functions
\begin{align}
	\label{eq:pgcn:zeroimpulsepart}
	\varphi^{(1)}(\lambda) &= \begin{cases} 1 & \text{if } \lambda=0, \\ 0 & \text{else,} \end{cases} &&\text{(zero impulse part)}\\
	\label{eq:pgcn:lowrankpinvpart}
	\varphi^{(2)}(\lambda) &= \begin{cases} \frac{\lambda_1}{\lambda} & \text{if } 0 < \lambda \leq \lambda_r, \\ 0 & \text{else,} \end{cases} && \text{(low-rank pseudoinverse part)} \\
	\label{eq:pgcn:highpasspart}
	\varphi^{(3)}(\lambda) &= \begin{cases} \lambda_1 & \text{if } \lambda > \lambda_r, \\ 0 & \text{else.} \end{cases} && \text{(high-pass part)}
\end{align}
This means that the individual filter functions \eqref{eq:learnedfilter} are equal to \eqref{eq:filter3d} with $\alpha=W_{ij}^{(1)}$, $\beta=W_{ij}^{(2)}$, and $\gamma=W_{ij}^{(3)}$.
The corresponding convolutional matrices can be set up as
\begin{equation}
	\label{eq:pgcn:threeconvmatrices}
	\mathcal{K}^{(1)} = u_0 u_0^T, \quad \mathcal{K}^{(2)} = \lambda_1 U_r \Lambda_r^{-1} U_r^T, \quad \mathcal{K}^{(3)} = \lambda_1 \left( I - u_0 u_0^T - U_r U_r^T \right).
\end{equation}
This feature map can now be embedded in a classical GNN architecture.
We follow the traditional GCN setup for semi-supervised classification in that we use two layers each consisting of the convolutional feature map with an added bias and ReLU activation between the layers. The numbers of channels before and after each layer are given by the feature dimension $d$, the \emph{hidden width} hyperparameter $h$, and the number of classes $m$.
Let $X^{(0)} \in \mathbb{R}^{n \times d}$ be the input feature matrix.
Extending the notation by an index for the layer, we get the propagation scheme
\begin{equation}
\label{eq:neuralnetwork}
X^{(1)} = \sigma\left( \sum_{k=1}^{3} \mathcal{K}^{(k)} X^{(0)} W^{(1,k)} + b^{(1)} \right),
\quad
X^{(2)} = \sum_{k=1}^{3} \mathcal{K}^{(k)} X^{(1)} W^{(2,k)} + b^{(2)}.
\end{equation}
Here $\sigma$ is the ReLU function applied element-wise and $W^{(1,k)} \in \mathbb{R}^{d \times h}$, $b^{(1)} \in \mathbb{R}^h$, $W^{(2,k)} \in \mathbb{R}^{h \times m}$, and $b^{(2)} \in \mathbb{R}^m$ are the trainable network parameters ($k=1,\ldots,K$). The addition of the biases $b^{(l)}$ to the matrices $\sum_{k=1}^K \mathcal{K}^{(k)} X^{(l-1)} W^{(l,k)}$ is understood row-wise. i.e., $b^{(l)}$ is added as a row vector to each row of the former matrix ($l=1,2$).

\subsection{Computational aspects}
\label{sec:pgcn:computational}

\paragraph{Setup.}
In order to assemble the convolutional matrices, we only need to compute the $r+1$ smallest eigenvalues of $\mathcal{L}_{\mathrm{sym}}$. This can be done by efficiently computing the largest eigenvalues of the \emph{signless Laplacian} via the state-of-the-art Krylov-Schur method \citep{stewart02}. Since the eigenvector to $\lambda_0=0$ is known a priori via \eqref{eq:firsteigenvector}, we can use Wielandt deflation to remove the eigengap and significantly accelerate the method \citep[Chapter 4.2]{saad11}. Put together, the system matrix of the eigenvalue computation is
\[
	2I-\mathcal{L}_{\mathrm{sym}}-2 u_0 u_0^T = I+D^{-1/2}AD^{-1/2}-2 u_0 u_0^T.
\]
If $\mu_0 \geq \ldots \geq \mu_{r-1}$ are the $r$ largest eigenvalue of that matrix, then the non-zero eigenvalues of $\mathcal{L}_{\mathrm{sym}}$ can be recovered via $\lambda_i = 2-\mu_{i-1}$ for $i = 1,\ldots,r$. 
The corresponding eigenvectors are the same. 
This way, the asymptotic setup cost amounts to the number of Krylov iterations times the cost of one matrix-vector product, which is $\mathcal{O}(n^2)$ in the worst case but may be significantly less if we are able to exploit any special problem-dependent structure. The number of required iterations, on the other hand, depends on the desired rank $r$.

\paragraph{Asymptotic cost of layer operations.}

Similar to $\mathcal{L}_{\mathrm{sym}}^\dagger$, the convolutional matrices $\mathcal{K}^{(k)}$ from \eqref{eq:pgcn:threeconvmatrices} will in general not be sparse. However, we do not have to store and apply the full matrices, but rather exploit their low rank by keeping them in their factorized form 
and computing the feature map \eqref{eq:pgcn:featuremap} via
\begin{equation}
\label{eq:efficientfeaturemap}
\begin{aligned}
	Y &= u_0 \left((u_0^T X) W^{(1)} - \lambda_1 u_0^T (X W^{(3)})\right) \\ 
	& \qquad + \; \lambda_1 U_r \left( \Lambda_r^{-1} (U_r^T X) W^{(2)} - U_r^T (X W^{(3)})\right) \; + \; X W^{(3)}.
\end{aligned}
\end{equation}
This way, the asymptotic cost of a single feature map 
is $\mathcal{O}(N_0 nr)$, where $N_0$ is the number of input features. Note that in general, multiplications with the dense adjacency matrix are already in $\mathcal{O}(n^2)$.

\paragraph{Choice of rank.}
As stated in Section~\ref{sec:framework:lowrank}, the target rank $r$ should be chosen roughly as the number of informative eigenvalues.
However, higher ranks may have additional benefits.
Typically, the rank choice for low-rank approximations comes down to a trade-off between accuracy and runtime.
An alternative is to view $r$ as a meta-parameter to be determined via cross validation.
We investigate its influence in Section~\ref{sec:rankdependency}.
Note that it is very common for GNNs to depend on a few parameters that control some level of approximation.

\section{Fast setup in special cases}

Certain application settings allow us to exploit intrinsic structure of the adjacency matrix to speed up the required eigenvalue computations, as we discuss now.

\subsection{Three-dimensional point clouds}
\label{sec:pgcn:nfft}

One source of dense graphs are collections of points in three-dimensional space, called point clouds, which may be produced by LiDAR scans or other applications. 
The task of semi-supervised point classification occurs when a segmentation of such a scan is desired based on labels assigned manually to a few points.
Other important tasks associated with this type of data are supervised point cloud segmentation and classification, which both revolve around transferring knowledge from fully labeled point clouds to new unlabeled point clouds.

One popular approach to working with this data, especially in robotics, is to turn the point cloud into a graph, e.g., by a $k$-Nearest Neighbor (KNN) setup \citep{nguyen13pointclouds, golovinskiy09}.
A promising alternative is to form a fully connected graph with Gaussian edge weights, leading to an adjacency matrix with entries
\begin{equation}
\label{eq:pgcn:gaussian}
A_{ij} = \exp\left(\frac{-\|x_i-x_j\|^2}{\sigma^2}\right)
\end{equation}
for all $i\neq j$, where the $x_i \in \mathbb{R}^3$ denote the point coordinates and $\sigma > 0$ is a localization parameter. This graph setup is often used for Spectral Clustering \citep{ng01spectralclustering}. A smaller value for $\sigma$ leads to a sparser graph. However, it may be beneficial for the graph to incorporate more non-local information by means of a larger value for $\sigma$, which leads to a dense graph. 

In this setting, the pseudoinverse assembly can be accelerated considerably.
The smallest Laplacian eigenvalues and corresponding eigenvectors can be approximated accurately and efficiently by exploiting a fast summation scheme based on the Non-Equispaced Fast Fourier Transform \citep[NFFT; see][]{alfke18}. This method has the remarkable property that it avoids assembling the $n^2$ adjacency entries altogether and that its computational effort for computing a small number of eigenvalues only scales linearly in $n$ instead of the usual quadratic behaviour. By combining the NFFT algorithm with our low-rank approximation scheme, 
the computational effort is significantly reduced, especially for large $n$.
Even though the amount of similarity information we look at is in $\mathcal{O}(n^2)$ and we do not discard any structure, the total complexity of our method with constant $r$ scales only like $\mathcal{O}(n)$. For all details on the NFFT method, we refer to \citet{alfke18}. Note that the NFFT-based fast multiplication for point cloud data Laplacians is applicable beyond our proposed filter whenever such a graph Laplacian matrix vector product is required.

\subsection{Hypergraphs for categorical data}
\label{sec:hypergraphs}

While the edges of traditional graphs connect exactly two nodes with each other, the \emph{hyperedges} of a hypergraph may connect any number of nodes \citep{bretto13}. Hypergraphs are most commonly described by their incidence matrix $H \in \{0,1\}^{n \times |E|}$, where $|E|$ denotes the number of hyperedges and $H_{ie} = 1$ if and only if node $i$ is a member of hyperedge $e$. Optional hyperedge weights can be given in a diagonal matrix $W_E = \diag(w_1,\ldots,w_{|E|})$. 
In addition to the node degree matrix $D \in \mathbb{R}^{n \times n}$ with entries $D_{ii} = \sum_{e=1}^{|E|} H_{ie} w_e$, we also set up the hyperedge degree matrix $B \in \mathbb{R}^{|E|\times|E|}$ with entries $B_{ee} = \sum_{i=1}^n H_{ie}$.

\paragraph{The hypergraph Laplacian operator.}
The hypergraph Laplacian can be defined in multiple ways.
We will use the linear definition introduced by \citet{zhou06}, given as 
\begin{equation}
\label{eq:hypergraph:laplacian}
\mathcal{L}_{\mathrm{sym}} = I - D^{-1/2} H W_E B^{-1} H^T D^{-1/2}.
\end{equation}
This is identical to the graph Laplacian of a classical graph with weighted adjacency matrix $H W_E B^{-1} H^T$, which is referred to as the \emph{clique expansion} of the hypergraph. This graph contains a specific set of loops and is in most applications dense or even fully connected.
As a consequence, naive computations with $\mathcal{L}_{\mathrm{sym}}$ may become quite expensive.
This problem also affects Hypergraph Neural Networks \citep{feng19}, which essentially apply the standard GCN architecture (including self loops) to the clique expansion graph.

\paragraph{Efficient techniques for the special case.}
One application of hypergraphs is categorical data where each sample is described by a few categorical attributes. We can simply construct one hyperedge for each possible value of an attribute, connecting all the samples which share that particular attribute value. This leads to the number of hyperedges being significantly smaller than the number of samples, $|E| \ll n$.
For other automatically generated hypergraphs, there is precedence for the benefits of generating fewer, larger hyperedges as well \citep{purkait17}.

In this special case, the Laplacian definition directly exhibits a useful structure. The matrix subtracted from the identity in \eqref{eq:hypergraph:laplacian} has rank $|E|$, so $\mathcal{L}_{\mathrm{sym}}$ is a linear combination of the identity and a low-rank matrix and can be written as
\begin{equation}
\label{eq:hypergraph:lowrankstructure}
\mathcal{L}_{\mathrm{sym}} = I - \tilde{H} \tilde{H}^T, \qquad\qquad \tilde{H} = D^{-1/2} H W_E^{1/2} B^{-1/2} \in \mathbb{R}^{n \times |E|}.
\end{equation}
Assume that $\tilde{H}$ has full rank $|E|$ and that its \emph{thin} singular value decomposition is given by
\begin{equation}
\tilde{H} = \tilde{U} \tilde{\Sigma} \tilde{V}^T,
\end{equation}
where $\tilde{U} \in \mathbb{R}^{n \times |E|}$ and $\tilde{V} \in \mathbb{R}^{|E| \times |E|}$ have orthogonal columns and $\tilde{\Sigma} \in \mathbb{R}^{|E| \times |E|}$ holds the singular values on its diagonal. Then $n-|E|$ eigenvalues of $\mathcal{L}_{\mathrm{sym}}$ are 1 and the remaining eigenvalues are given by $\tilde{\Lambda} = I - \tilde{\Sigma}^2$. Consequently, the exact pseudoinverse of $\mathcal{L}_{\mathrm{sym}}$ is the identity plus a matrix of rank $|E|$ and has the structure
\begin{equation}
\label{eq:hypergraph:efficientpseudoinverse}
\mathcal{L}_{\mathrm{sym}}^\dagger = I + \tilde{U} 
(\tilde{\Lambda}^\dagger - I)
\tilde{U}^T,
\quad
\tilde{\Lambda}^\dagger = 
\diag\left(0,\ \frac{1}{\lambda_1},\ \ldots,\ \frac{1}{\lambda_{|E|-1}}\right).
\end{equation}
For our low-rank approach, this means that we cannot choose $r > |E|-1$ because that would require singling out a few eigenvectors to the eigenvalue 1, which are indistinguishable. However, computing the first $r+1 \leq |E|$ eigenvalues of the hypergraph Laplacian becomes much cheaper, since we only need the singular value decomposition of $\tilde{H}$ or equivalently the eigenvalue decomposition of the $|E|\times|E|$ matrix $\tilde{H}^T \tilde{H}$. Thus the setup cost is significantly reduced. 
Furthermore, we can recreate the full-rank filter \eqref{eq:filter2d} within the low-rank setting of \eqref{eq:filter3d} by fixing $r=|E|-1$ and $\gamma=\beta$.

\section{Experimental results}
\label{sec:experiments}

\subsection{Network architecture and training setup}
\label{sec:experiments:setup}

Our code is available online,\footnote{\url{https://github.com/dominikalfke/PinvGCN}} implemented in Python using PyTorch and PyTorch Geometric \citep{fey19torchgeometric}. 
The input features of each dataset are propagated through the network architecture \eqref{eq:neuralnetwork} with the hidden width set to $h=32$ in all experiments.
In the first layer, the products $\mathcal{K}^{(k)} X^{(0)}$ are precomputed as proposed by \citet{chen18}, and in the second layer we employ the efficient scheme from \eqref{eq:efficientfeaturemap}.
Afterwards, the rows of the output $X^{(2)}$ are transformed via the softmax function to gain the predicted class probabilities for each sample. The parameters are trained using the average cross entropy loss of the predicted probabilities for the true class of the training samples. We use the Adam optimizer \citep{adam} with learning rate 0.01. During training, we use a dropout rate of 0.5 between the layers. For the weight matrices $W^{(l,k)}$, we use Glorot initialization \citep{glorot10} and a weight decay factor of 0.0005, while for the bias vectors $b^{(l)}$, we use zero initialization and no weight decay.

For each run, we set a fixed seed for random number generation, build the model, run 500 training epochs, and finally evaluate the classification accuracy on the non-training samples. We generally perform 100 of these runs for each experimental setting and report averages and standard deviations. Only in a few slow baseline experiments did we reduce the number of runs to 10. 
All experiments are run on a laptop with an Intel 
Core i7 processor and an NVIDIA GeForce RTX 2060.

\subsection{Baselines}

We refer to our own method as PinvGCN in plots and tables, where we compare it against the following methods on graphs: 
\begin{itemize}
	\item GCN \citep{kipf17}, using the implementation from PyTorch Geometric.
	
	\item GraphSAGE \citep{hamilton17}, using the implementation from PyTorch Geometric with mean aggregation. We do not use neighbor sampling since that is designed for node batches in large datasets and it did not improve the results in any of our tests.
	
	\item GDC \citep{klicpera19}, using the implementation from PyTorch Geometric with $\alpha=0.05$ and top-64 sparsification. Since the datasets where too large for exact matrix inversion, we needed to use the ``inexact'' version,  which is not supported in the original code published with the paper.\footnote{\url{https://github.com/klicperajo/gdc}} 

	\item ARMA \citep{bianchi19}, using the implementation from PyTorch Geometric with parameters $K=3$ and $T=2$.
\end{itemize}
On hypergraphs we additionally tested the following two methods:
\begin{itemize}
	\item HGNN* \citep{feng19}, which we mark with an asterisk because we use our own implementation that exploits the structure of \eqref{eq:hypergraph:lowrankstructure} in a similar way to our own method, significantly speeding up the runtime without changing the output.
	
	\item HyperGCN \citep{yadati19}, using the code published with the paper.\footnote{\url{https://github.com/malllabiisc/HyperGCN}} Since the fast and non-fast variant yield the same accuracy in our experiments, we only employ ``FastHyperGCN with mediators''.
\end{itemize}

\subsection{Semi-supervised point classification in 3D point clouds}
\label{sec:experiments:oakland}

In order to illustrate our method's superior performance on very dense graphs, we employ two 3D point clouds as described in Section~\ref{sec:pgcn:nfft}. We use a part of a subset of the popular Oakland dataset as used by \citet{munoz09oakland}. 
We obtained the subset of the Oakland dataset from the project website.\footnote{\url{http://www.cs.cmu.edu/~vmr/datasets/oakland_3d/cvpr09/doc/}}
The dataset consists of multiple point clouds, two of which are used for training and validation in the original setting. Since their original usage is different from our own training splits, we only refer to the clouds as Oakland 1 (original training cloud) and Oakland 2 (original validation cloud). The remaining test clouds are unused. This data is usually used for 3D point segmentation, where the task is to transfer knowledge from one cloud to other clouds \citep{nguyen13pointclouds}. Instead, we look at the two clouds independently and perform 5-class semi-supervised point classification on each of them by splitting each cloud into its own sets of training points and test points. We use 100 training points from each of the  five classes. For the input feature matrix $X^{(0)}$, we use the original 3D coordinates. Dataset information is given in Table~\ref{tbl:oaklandinfo}, where the diameter denotes the maximum Euclidean distance between two points in the cloud.

\begin{table}
	\caption{Information on the Oakland point clouds}
	\label{tbl:oaklandinfo}
	
	\begin{center}
		\begin{tabular}{@{}lcccccc}
			\toprule
			\multirow{2}{*}{Name} & \multirow{2}{*}{Nodes} & \multirow{2}{*}{Classes} & \multirow{2}{*}{Label rate} & \multirow{2}{*}{Diameter} & \multicolumn{2}{c}{Eigengap $\lambda_1$} \\\cmidrule(lr){6-7}
			&&&&& $\sigma=10$ & $\sigma=100$ \\
			\midrule
			Oakland 1 & 36\,932 & 5 & 1.35 \% & 112.1 & 0.084 & 0.929 \\
			Oakland 2 & 91\,515 & 5 & 0.55 \% & 126.9 & 0.002 & 0.872 \\
			\bottomrule
		\end{tabular}
	\end{center}
\end{table}

For eigenvalue computation, we use the \texttt{fastadj} Python implementation\footnote{\url{https://github.com/dominikalfke/FastAdjacency}} of the NFFT-based fast summation scheme \citep{alfke18} with default parameters, combined with the Krylov-Schur algorithm with tolerance $10^{-3}$. These settings are chosen to give fast, rough approximations of the eigenvalues because we found that higher quality did not have any influence on the PinvGCN results.

\begin{table}
	\caption{Average results on Oakland datasets}
	\label{tbl:oakland}
	\begin{center}
		\begin{tabular}{llccccc}
			\toprule
			\multicolumn{2}{l}{\multirow{2}{*}{Method}} & \multirow{2}{*}{$\sigma$} & \multicolumn{2}{c}{Oakland 1} & \multicolumn{2}{c}{Oakland 2} \\ 
			\cmidrule(lr){4-5} \cmidrule(lr){6-7}
			& & & Time & Accuracy ($\pm$ SD) & Time & Accuracy ($\pm$ SD) \\
			\midrule
			
			\multirow{5}{*}{\rotatebox[origin=c]{90}{$10$-NN}}
			& \multirow{2}{*}{GCN}
			& 10
			& 4.23 s &  58.16 \% ($\pm$  4.39)    & 5.93 s &  68.70 \% ($\pm$  18.28)    \\
			&& 100
			& 4.24 s &  58.42 \% ($\pm$  4.10)    & 5.91 s &  68.64 \% ($\pm$  3.28)    \\
			& GraphSAGE & --
			& 2.61 s &  56.22 \% ($\pm$  5.57)    & 3.50 s &  71.32 \% ($\pm$  4.26)    \\
			& GDC & --
			& 112.08 s & 48.61 \% ($\pm$ 8.50)   & 1160.33 s & 63.25 \% ($\pm$ 25.15)   \\
			& ARMA & 100
			& 29.23 s &  49.63 \% ($\pm$  2.55)    & 451.49 s &  64.29 \% ($\pm$  4.41)    \\
			\midrule
			
			\multirow{3}{*}{\rotatebox[origin=c]{90}{$100$-NN}}
			& \multirow{2}{*}{GCN}
			& 10
			& 17.99 s &  35.44 \% ($\pm$  11.87)    & 44.04 s &  66.47 \% ($\pm$  20.54)    \\
			&& 100
			& 17.67 s &  54.17 \% ($\pm$  6.49)    & 43.66 s & 70.89 \% ($\pm$ 3.27)   \\
			& GraphSAGE & --
			& 6.63 s &  59.34 \% ($\pm$  4.94)    & 13.76 s &  70.55 \% ($\pm$  3.98)    \\
			& GDC & --
			& 3446.95 s & 37.84 \% ($\pm$ 5.09)   & \multicolumn{2}{c}{\multirow{2}{*}{Out of memory}}
			\\
			& ARMA & 100
			& 76.02 s &  56.04 \% ($\pm$  4.51)    & & 
			\\
			\midrule

			\multicolumn{2}{l}{\multirow{2}{*}{PinvGCN, $r=10$}}
			& 10
			& 6.70 s & 84.64 \% ($\pm$ 4.56)   & 14.86 s & 74.26 \% ($\pm$ 6.39)   \\
			&& 100
			& 5.79 s & 92.58 \% ($\pm$ 1.42)   & 10.82 s & 93.25 \% ($\pm$ 1.02)   \vspace{0.2em}\\
			
			\multicolumn{2}{l}{\multirow{2}{*}{PinvGCN, $r=30$}}
			& 10
			& 11.02 s & 81.58 \% ($\pm$ 4.64)   & 24.62 s & 74.18 \% ($\pm$ 6.61)   \\
			&& 100
			& 11.08 s & 93.13 \% ($\pm$ 1.68)   & 24.03 s & 94.91 \% ($\pm$ 0.93)   \vspace{0.2em}\\
			
			\multicolumn{2}{l}{\multirow{2}{*}{PinvGCN, $r=100$}}
			& 10
			& 29.80 s & 81.70 \% ($\pm$ 5.02)   & 69.54 s & 73.29 \% ($\pm$ 6.18)   \vspace{0.2em}\\
			&& 100
			& 28.93 s & 93.50 \% ($\pm$ 1.83)   & 67.70 s & 95.38 \% ($\pm$ 0.90)   \vspace{0.2em}\\

			\bottomrule
		\end{tabular}
	\end{center}
\end{table}

We conduct experiments for $\sigma \in \{10,100\}$ in the Gaussian function \eqref{eq:pgcn:gaussian}, where the larger value amounts to a stronger inclusion of non-local information.
Our experimental results are shown in Table~\ref{tbl:oakland}.
Since the baselines are not designed for such dense graphs and would have exploding time and memory requirements on the fully connected graph, we only employ them on a $k$-NN subgraph of the $k$ nearest neighbors, where $k$ is either 10 or 100. 
Note that it would be possible to employ the GCN baseline on the fully connected graph by utilizing the same NFFT-based fast summation, but providing such an implementation was beyond the scope of our experiments.

To summarize the results, our method 
produces accurate predictions in reasonable time, comfortably outperforming all baselines.
The fact that our accuracy is substantially better for $\sigma = 100$ shows that our method greatly profits from non-local information in non-sparse Laplacians. As we see in Table~\ref{tbl:oaklandinfo}, adding non-local information via a larger value of $\sigma$ results in increasing $\lambda_1$, confirming the connection between spatial and spectral properties.

\subsection{Hypergraph datasets from categorical attributes}
\label{sec:exp:hypergraphs}

We finally employ our method on three hypergraphs based on the \emph{Mushroom} and \emph{Covertype} datasets from the UCI machine learning repository \citep{UCIwebsite}, which are common benchmarks for semi-supervised classification on hypergraphs \citep{hein13,yadati19}.

\begin{table}
	\caption{Hypergraph datasets}
	\label{tbl:exp:hypergraphinfo}
	\begin{center}
		\begin{tabular}{@{}lcccc@{}} 
			\toprule
			Name & $n$ & $|E|$ & Classes & $\lambda_1$ \\
			\midrule
Mushroom & 8\ 124 & 112 & 2 & 0.67 \\
			Covertype45 & 12\ 240 & 104 & 2 & 0.58 \\
			Covertype67 & 37\ 877 & 125 & 2 & 0.59 \\
			\bottomrule
		\end{tabular}
	\end{center}
\end{table}

The Mushroom dataset\footnote{\url{https://archive.ics.uci.edu/ml/datasets/Mushroom}} contains 8124 samples from two classes, described by 21 categorical attributes (ignoring one with missing values). For each attribute, we create as many hyperedges as there are attribute values present, where each hyperedge connects all samples with a specific value. 

The Covertype dataset\footnote{\url{https://archive.ics.uci.edu/ml/datasets/Covertype}} contains 581012 samples from 7 classes, described by 10 continuous and 44 binary attributes. We follow the setup process from \citep{hein13}, dividing the value range of each continuous attribute into 10 equally sized bins and creating hyperedges that connect all samples with values in the same bin. For each binary attribute, we only create one hyperedge for those samples with a ``true'' value. All hyperedges have weight $w_e=1$. Afterwards, we create two subhypergraphs by using only samples from classes 4 and 5, or 6 and 7. Because we remove all hyperedges with less than two nodes, the resulting hypergraphs have less than the original 144 hyperedges.

As in \citep{yadati19}, we use the hypergraph incidence matrix as input features for all three data sets, $X^{(0)} = H$.
Table~\ref{tbl:exp:hypergraphinfo} gives the full hypergraph specifications as well as their smallest nonzero Laplacian eigenvalue.

\begin{table}
	\caption{Results for UCI categorical dataset with 10 training samples per class}
	\label{tbl:exp:hypergraphs}
	
	\begin{subtable}{\textwidth}
		\begin{center}
			\begin{tabular}{@{}clcc@{}}
				\toprule
				\multicolumn{2}{l}{\multirow{2}{*}{Method}} & \multicolumn{2}{c}{Mushroom} \\
				\cmidrule(lr){3-4}
				&& Runtime & Accuracy ($\pm$ SD)  \\
				\midrule
				\multirow{4}{*}{\,\,\,\,\rotatebox[origin=c]{90}{Sparsified} \rotatebox[origin=c]{90}{graph}}
				& GCN & 2.93 s &  91.46 \% ($\pm$  2.99) \\   & GraphSAGE & 2.61 s &  92.23 \% ($\pm$  3.71) \\   & GDC & 31.98 s &  92.24 \% ($\pm$  3.17) \\  & ARMA & 8.16 s &  \textbf{92.36 \%} ($\pm$  3.19) \\  \multicolumn{2}{l}{HGNN*} & 0.61 s & 83.38 \% ($\pm$ 11.23) \\  \multicolumn{2}{l}{FastHyperGCN} & 2.93 s & 76.13 \% ($\pm$ 16.38) \\  
				\multicolumn{2}{l}{PinvGCN, $r=|E|-1$}
				& 3.22 s & 91.38 \% ($\pm$ 3.92)   \\
				\multicolumn{2}{l}{PinvGCN without high-pass}
				& 2.77 s & 91.58 \% ($\pm$ 3.33)   \\

				\bottomrule
			\end{tabular}
		\end{center}
	\end{subtable}
	\begin{subtable}{\textwidth}
		\begin{center}
			\begin{tabular}{@{}clcccc@{}}
				\toprule
				\multicolumn{2}{l}{\multirow{2}{*}{Method}} &  \multicolumn{2}{c}{Covertype45} & \multicolumn{2}{c}{Covertype67} \\
				\cmidrule(lr){3-4} \cmidrule(lr){5-6}
				&& Runtime & Accuracy ($\pm$ SD)  & Runtime & Accuracy ($\pm$ SD)  \\
				\midrule
				
				\multirow{4}{*}{\,\,\,\,\rotatebox[origin=c]{90}{Sparsified} \rotatebox[origin=c]{90}{graph}}
				& GCN 
				& 1.49 s & 96.68 \% ($\pm$ 2.43)   & 2.80 s & 90.01 \% ($\pm$ 2.63)   \\
				& GraphSAGE 
				& 1.30 s & 97.80 \% ($\pm$ 1.76)   & 3.14 s & 92.91 \% ($\pm$ 2.33)   \\
				& GDC 
				& 138.49 s & 98.16 \% ($\pm$ 1.76)   & 417.18 s & 94.20 \% ($\pm$ 2.72)   \\
				& ARMA 
				& 9.79 s & 96.96 \% ($\pm$ 3.39)   & 30.39 s & 93.53 \% ($\pm$ 2.56)   \\
				\multicolumn{2}{l}{HGNN*} 
				& 0.62 s & 94.83 \% ($\pm$ 15.75)   & 1.09 s & 91.07 \% ($\pm$ 10.78)   \\
				\multicolumn{2}{l}{FastHyperGCN} 
				& 3.51 s & 89.81 \% ($\pm$ 8.72)   & 9.00 s & 80.57 \% ($\pm$ 13.30)   \\

				\multicolumn{2}{l}{PinvGCN, $r=|E|-1$}
				& 3.33 s & 99.58 \% ($\pm$ 0.71)   & 3.55 s & 96.33 \% ($\pm$ 1.41)   \\
				\multicolumn{2}{l}{PinvGCN without high-pass}
				& 2.71 s & 99.56 \% ($\pm$ 0.81)   & 3.02 s & 96.27 \% ($\pm$ 1.43)   \\
				\bottomrule
			\end{tabular}
		\end{center}
	\end{subtable}
\end{table}

Classification results on all three datasets are listed in Table~\ref{tbl:exp:hypergraphs}. We only employ our Pseudoinverse GCN with the maximum rank $r=|E|-1$, as will be supported in Section~\ref{sec:rankdependency}. For the graph-based methods, we use KNN sparsification of the clique expansion graph, so each node is connected to the $k=10$ other nodes with highest shared hyperedge membership. 
On the Mushroom hypergraph, our results are in the same order of magnitude as the graph baselines, which is better than the other hypergraph methods, but the ARMA network gives superior results. The graph methods perform remarkably well considering the fact that this process of clique expansion sparsification has, to our knowledge, never been discussed in the hypergraph literature. On both Covertype datasets, however, our Pseudoinverse GCN yields the best accuracies.

In addition, we also emulated the full-rank pseudoinverse filter \eqref{eq:filter2d} via fixing $\beta=\gamma$ in \eqref{eq:filter3d} with $r=|E|-1$. This method is named \emph{PinvGCN without high-pass} in Table~\ref{tbl:exp:hypergraphs}
and it produces comparable results. This shows that in the case that all informative eigenvalues can be computed, the high-pass part is no longer obligatory. However, manually removing this part also does not increase accuracy, so we recommend always using all three filter basis functions.

\begin{figure}
	\begin{center}
		\begin{tikzpicture}
		\begin{axis}[
		width=\linewidth,
		height=2cm,
		hide axis,
		xmin=0, xmax=1, ymin=0, ymax=1,
		font=\footnotesize,
		legend columns = -1,
		legend style={
			legend cell align=left, 
			at={(0.5,0.5)}, 
			anchor=center,
			/tikz/every even column/.append style={column sep=0.5cm}},
		xtick = {5,20,50},
		ymajorgrids,
		]
		
		\addlegendimage{darkgreen, mark=square, dashed, thick, mark size=3pt, mark options={solid}}
		\addlegendentry{GDC}
		
		\addlegendimage{red, mark=x, dotted, thick, mark size=3pt, mark options={solid}}
		\addlegendentry{ARMA}
		
		\addlegendimage{cyan, mark=triangle, dashdotted, thick, mark size=3pt, mark options={solid}}
		\addlegendentry{HGNN*}
		
		\addlegendimage{blue, mark=o, thick, mark size=3pt, mark options={solid}}
		\addlegendentry{PinvGCN, $r=|E|-1$}
		
		\end{axis}
		\end{tikzpicture}
	\end{center}
	
	\def\w{2.7cm}
	\def\h{5cm}
	\begin{subfigure}{0.36\textwidth}
		\begin{tikzpicture}
		\begin{axis}[
		width=\w,
		height = \h,
		scale only axis,
xlabel = {\vphantom{Training samples per class}},
		ylabel = {\footnotesize Misclassification rate (\%)},
		every tick label/.append style={font=\footnotesize},
		ymode = log,
		xtick = {5,10,20,50},
		xticklabels = {5,,20,50},
		log ticks with fixed point,
		ytick={0.2, 0.1, 0.05, 0.025},
		yticklabels={20, 10, 5, 2.5},
		xmajorgrids,
		ymajorgrids
		]
		
\addplot[darkgreen, mark=square, dashed, thick, mark size=3pt, mark options={solid}] coordinates { (5, 1.2781e-01) (10, 7.7615e-02) (20, 5.0675e-02) (50, 2.4903e-02) };
		
\addplot[cyan, mark=triangle, dashdotted, thick, mark size=3pt, mark options={solid}] coordinates { (5, 1.4535e-01) (10, 1.2106e-01) (20, 1.1441e-01) (50, 1.1126e-01) };
		
\addplot[red, mark=x, dotted, thick, mark size=3pt, mark options={solid}] coordinates { (5, 1.2098e-01) (10, 7.6091e-02) (20, 5.1336e-02) (50, 2.5904e-02) };

		\addplot[blue,mark=o, thick, mark size=3pt, mark options={solid}] coordinates { (5, 1.2429e-01) (10, 8.6156e-02) (20, 5.8137e-02) (50, 3.1473e-02) };
		
		\end{axis}
		\end{tikzpicture}
		\caption{Mushroom}
	\end{subfigure}\begin{subfigure}{0.36\textwidth}
		\begin{tikzpicture}
		\begin{axis}[
		width=\w,
		height = \h,
		scale only axis,
		xlabel = {\footnotesize Training samples per class},
every tick label/.append style={font=\footnotesize},
		ymode = log,
		xtick = {5,10,20,50},
		xticklabels = {5,,20,50},
		log ticks with fixed point,
		ytick={0.1, 0.01, 0.001, 0.0001, 0.00001},
		yticklabels={10, 1, 0.1, 0.01, 0.001},
		xmajorgrids,
		ymajorgrids,
		]

\addplot[darkgreen, mark=square, dashed, thick, mark size=3pt, mark options={solid}] coordinates { (5, 6.8200e-02) (10, 1.6619e-02) (20, 5.0254e-03) (50, 2.3501e-03) };
		
\addplot[cyan, mark=triangle, dashdotted, thick, mark size=3pt, mark options={solid}] coordinates { (5, 5.6742e-02) (10, 1.0642e-02) (20, 4.5943e-03) (50, 2.0231e-03) };
		
\addplot[red, mark=x, dotted, thick, mark size=3pt, mark options={solid}] coordinates { (5, 8.0349e-02) (10, 2.4019e-02) (20, 1.1684e-02) (50, 7.1639e-03) };

\addplot[blue, mark=o, thick, mark size=3pt, mark options={solid}] coordinates {(5, 4.7453e-02) (10, 4.1571e-03) (20, 5.4426e-04) (50, 2.4712e-06)};
		
		\end{axis}
		\end{tikzpicture}
		\caption{Covertype45}
	\end{subfigure}\begin{subfigure}{0.28\textwidth}
		\begin{tikzpicture}
		\begin{axis}[
		width=\w,
		height = \h,
		scale only axis,
xlabel = {\vphantom{Training samples per class}},
every tick label/.append style={font=\footnotesize},
		ymode = log,
		xtick = {5,10,20,50},
		xticklabels = {5,,20,50},
		log ticks with fixed point,
		ytick={0.09, 0.03, 0.01, 0.001},
		yticklabels={9, 3, 1},
		xmajorgrids,
		ymajorgrids,
		]

\addplot[darkgreen, mark=square, dashed, thick, mark size=3pt, mark options={solid}] coordinates { (5, 1.0085e-01) (10, 4.8963e-02) (20, 2.4195e-02) (50, 8.2492e-03) };
		
\addplot[cyan, mark=triangle, dashdotted, thick, mark size=3pt, mark options={solid}] coordinates { (5, 7.4896e-02) (10, 5.6728e-02) (20, 4.3861e-02) (50, 3.0991e-02) };
		
\addplot[red, mark=x, dotted, thick, mark size=3pt, mark options={solid}] coordinates { (5, 1.2038e-01) (10, 6.4908e-02) (20, 3.6361e-02) (50, 1.7455e-02) };

\addplot[blue, mark=o, thick, mark size=3pt, mark options={solid}] coordinates {(5, 6.7881e-02) (10, 3.6710e-02) (20, 2.0059e-02) (50, 6.3528e-03)};
		
		\end{axis}
		\end{tikzpicture}
		\caption{Covertype67}
	\end{subfigure}
	\caption{Misclassification rate development over different split sizes for hypergraph datasets}
	\label{fig:experiments:hypergraphs}
\end{figure}
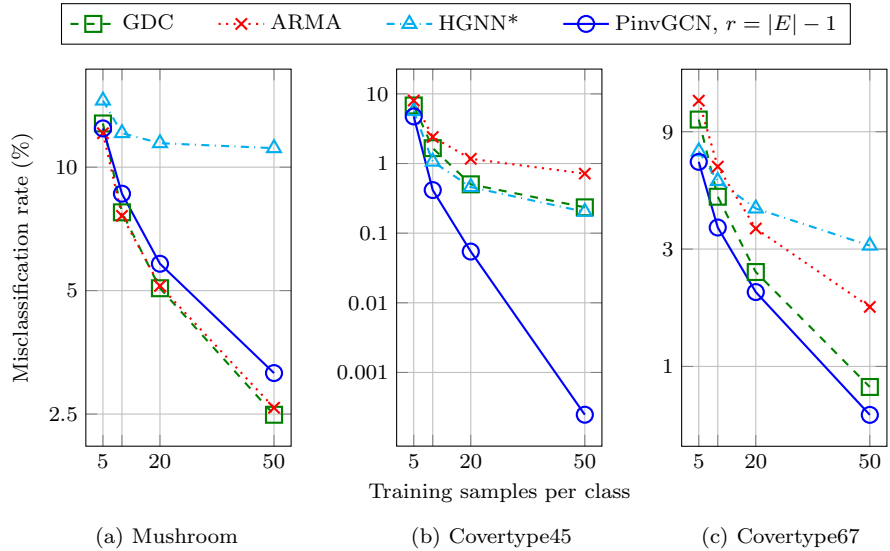

For further investigation, Figure~\ref{fig:experiments:hypergraphs} presents the dependency of the misclassification rate (i.e., the complement of the accuracy, here plotted logarithmically) on the number of random training samples. The plots confirm the results of Table~\ref{tbl:exp:hypergraphs} in principle. A remarkable finding is that our method has a significantly better convergence rate on the Covertype45 hypergraph.

\subsection{Limitations: Sparse graph datasets}
\label{sec:exp:sparse}

For comparison and transparency, we also apply our method to the standard citation networks Cora, Citeseer, and Pubmed. These graph datasets are typical benchmarks for GNNs. However, they all are examples of sparse graphs that are exactly the opposite of what our method is designed to handle. The main challenge of these examples for our method are the very small eigengaps. Moreover, the eigenvalue zero often has multiplicity larger than one, which requires putting more effort into the eigenvalue computation.
Network statistics and full results are given in Table~\ref{tbl:citation}.
Unsurprisingly, our method by itself produces poor results on these datasets and it does not come close to the performance of a simple GCN \citep{kipf17}. 

\begin{figure}
\begin{center}
\begin{tikzpicture}
	\begin{axis}[
		width=\linewidth,
		height=2cm,
		hide axis,
		xmin=0, xmax=1, ymin=0, ymax=1,
		font=\footnotesize,
		legend columns = -1,
		legend style={
			legend cell align=left, 
			at={(0.5,0.5)}, 
			anchor=center,
			/tikz/every even column/.append style={column sep=0.5cm}},
		xtick = {5,20,50},
		ymajorgrids,
		]
		
		\addlegendimage{blue, mark=square*, mark size=1pt, solid, thick}
		\addlegendentry{PinvGCN, rank $200$}
		
		\addlegendimage{red, mark=diamond*, mark size=1.5pt, solid, thick}
		\addlegendentry{PinvGCN, rank $500$}
		
		\addlegendimage{black, thick, dashed}
		\addlegendentry{GCN baseline}
		
	\end{axis}
\end{tikzpicture}

\begin{subfigure}{0.33\textwidth}
\begin{tikzpicture}
\begin{axis}[
		xlabel={Regularization rate $R$},
		ylabel={Accuracy (\%)},
		width=\textwidth,
		height = 5cm,
		every tick label/.append style={font=\footnotesize},
		xmode = log,
		xtick = {0.01, 0.1, 1, 10},
		log ticks with fixed point,
		xmin = 0.005, xmax = 20,
		ymin = 55, ymax=75,
		xmajorgrids,
		ymajorgrids,
		]

\addplot[blue, mark=square*, mark size=1pt, solid, thick] coordinates { (0.01, 61.670000) (0.02, 61.898000) (0.05, 64.249000) (0.1, 67.320000) (0.2, 68.335000) (0.5, 67.741000) (1, 65.864000) (2, 64.412000) (5, 63.132000) (10, 62.528000) };
		
\addplot[red, mark=diamond*, mark size=1.5pt, solid, thick] coordinates { (0.01, 63.237000) (0.02, 62.922000) (0.05, 66.430000) (0.1, 68.741000) (0.2, 68.987000) (0.5, 68.205000) (1, 67.291000) (2, 65.604000) (5, 63.571000) (10, 62.105000) };
		
		\addplot[black, dashed, thick, domain=0.005:20] {70.3};
		
	\end{axis}
\end{tikzpicture}
\caption{Citeseer}
\end{subfigure}\begin{subfigure}{0.33\textwidth}
\begin{tikzpicture}
\begin{axis}[
		xlabel={Regularization rate $R$},
		ylabel={Accuracy (\%)},
		width=\linewidth,
		height = 5cm,
		every tick label/.append style={font=\footnotesize},
		xmode = log,
		xtick = {0.01, 0.1, 1, 10},
		log ticks with fixed point,
		xmin = 0.005, xmax = 20,
		ymin = 65, ymax=85,
		xmajorgrids,
		ymajorgrids,
		]

\addplot[blue, mark=square*, mark size=1pt, solid, thick] coordinates { (0.01, 77.310000) (0.02, 77.889000) (0.05, 79.771000) (0.1, 79.923000) (0.2, 80.612000) (0.5, 80.507000) (1, 78.702000) (2, 75.208000) (5, 70.366000) (10, 67.848000) };
		
\addplot[red, mark=diamond*, mark size=1.5pt, solid, thick] coordinates { 
			(0.01, 77.833000) (0.02, 78.704000) (0.05, 80.919000) (0.1, 81.461000) (0.2, 82.336000) (0.5, 81.350000) (1, 78.226000) (2, 74.248000) (5, 69.776000) (10, 66.959000) };
		
		\addplot[black, dashed, thick] coordinates {(0.005, 80.18) (20, 80.18)};
		
	\end{axis}
\end{tikzpicture}
\caption{Cora}
\end{subfigure}\begin{subfigure}{0.33\textwidth}
\begin{tikzpicture}
	\begin{axis}[
		xlabel={Regularization rate $R$},
		ylabel={Accuracy (\%)},
		width=\linewidth,
		height = 5cm,
		every tick label/.append style={font=\footnotesize},
		xmode = log,
		xtick = {0.01, 0.1, 1, 10},
		log ticks with fixed point,
		xmin = 0.005, xmax = 20,
		ymin = 65, ymax=85,
		xmajorgrids,
		ymajorgrids,
		]

\addplot[blue, mark=square*, mark size=1pt, solid, thick] coordinates { (0.01, 76.539000) (0.02, 78.245000) (0.05, 78.303000) (0.1, 77.895000) (0.2, 75.419000) (0.5, 72.565000) (1, 70.528000) (2, 75.675000) (5, 79.228000) (10, 77.895000) };
		
\addplot[red, mark=diamond*, mark size=1.5pt, solid, thick] coordinates { (0.01, 78.084000) (0.02, 78.256000) (0.05, 76.349000) (0.1, 75.262000) (0.2, 75.485000) (0.5, 72.815000) (1, 72.504000) (2, 77.469000) (5, 78.180000) (10, 77.347000) };
		
		\addplot[black, dashed, thick, domain=0.005:20] {79.0};
		
	\end{axis}
\end{tikzpicture}
\caption{Pubmed}
\end{subfigure}
\end{center}
\caption{Results for graph datasets transformed into centroid hypergraphs with categorical regularization}
\label{fig:catreg}
\end{figure}
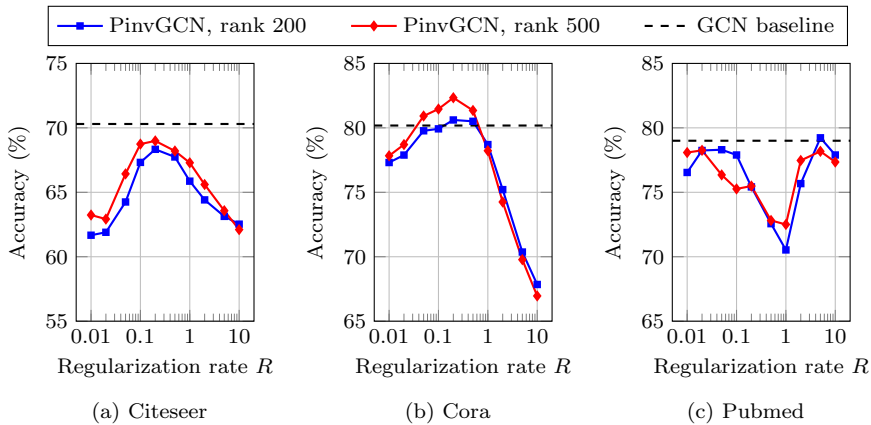

However, since these citation networks come with categorical node features, it is possible to use similar techniques as for the categorical hypergraphs in order to obtain a more dense structure. For a first step, we transform the graph into a hypergraph with one hyperedge per node, following \citet{feng19}. We then ``regularize'' the sparse graphs by adding one hyperedge for each column of the node feature matrix $X$, connecting all nodes with a nonzero feature value. Those feature-based hyperedges are weighted with a \emph{regularization factor} $R > 0$. Similarly to the hypergraphs from Section~\ref{sec:exp:hypergraphs}, the newly formed hypergraph Laplacian is considerably more dense and the eigengap grows monotonously with $R$. For each dataset, the results for various values of $R$ are visualized in Figure~\ref{fig:catreg} and the best obtained accuracy is listed in Table~\ref{tbl:citation}. The regularization helps our method catch up with state of the art, but in practice, this approach presents challenges in the high sensitivity and lack of a heuristic regarding $R$. It appears that for these citation networks, the sparse representation contains more useful clustering information than the node features. Our method is better suited to intrinsically dense data instead of relying on densification.

\begin{table}
	\caption{Results for sparse citation networks}
	\label{tbl:citation}
	\begin{center}
		\begin{tabular}{@{}lccc@{}}
			\toprule
			Method & Citeseer & Cora &  Pubmed \\
			\midrule
Number of nodes $n$ & 3327 & 2708 & 19717 \\
			Number of connected components $k$ & 438 & 78 & 1 \\
			Eigengap $\lambda_k$ & 0.0013 & 0.0036 & 0.0095 \\
			\midrule
GCN \citep[as reported in][]{kipf17} & 70.3 \% & 81.5 \% & 79.0 \% \\
PinvGCN, rank 50 & 61.14 \% & 70.79 \% & 70.76 \% \\
			PinvGCN, rank 200 & 60.90 \% & 71.15 \% & 71.53 \% \\
			PinvGCN, rank 500 & 61.62 \% & 71.41 \% & 71.58 \% \\
			PinvGCN, rank 500, best regularization & 68.33 \% & 82.34 \% & 78.26 \% \\
			
\bottomrule
		\end{tabular}
	\end{center}
\end{table}

\subsection{Rank dependency}
\label{sec:rankdependency}

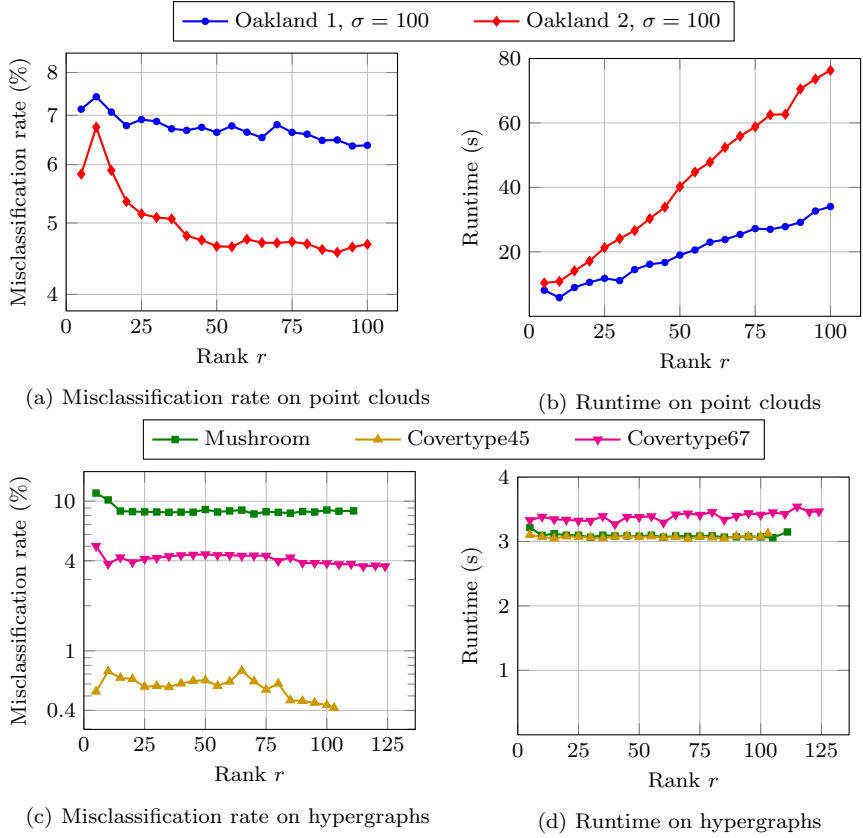
\begin{figure}
	\begin{center}
		\begin{tikzpicture}
		\begin{axis}[
		width=\linewidth,
		height=2cm,
		hide axis,
		xmin=0, xmax=1, ymin=0, ymax=1,
		font=\footnotesize,
		legend columns = -1,
		legend style={
			legend cell align=left, 
			at={(0.5,0.5)}, 
			anchor=center,
			/tikz/every even column/.append style={column sep=0.5cm}},
		xtick = {5,20,50},
		ymajorgrids,
		]
		
		\addlegendimage{blue, mark=*, mark size=1pt, solid, thick}
		\addlegendentry{Oakland 1, $\sigma=100$}
		
		\addlegendimage{red, mark=diamond*, mark size=1.5pt, solid, thick}
		\addlegendentry{Oakland 2, $\sigma=100$}
		
		\end{axis}
		\end{tikzpicture}
		
		\begin{subfigure}{0.5\textwidth}
			\begin{tikzpicture}
			\begin{axis}[
			xlabel={Rank $r$},
			ylabel={Misclassification rate (\%)},
			width=\linewidth,
			height = 5cm,
			every tick label/.append style={font=\footnotesize},
			ymode = log,
			xtick = {0,25,50,75,100,125},
			xmin = 0,
			log ticks with fixed point,
ytick = {8,7,6,5,4},
			ymin = 3.8, ymax = 8.5,
			xmajorgrids,
			ymajorgrids,
			]
			
\addplot[blue, mark=*, mark size=1pt, solid, thick] coordinates {  (5, 7.134689) (10, 7.417655) (15, 7.070844) (20, 6.777695) (25, 6.907499) (30, 6.866409) (35, 6.711325) (40, 6.678113) (45, 6.741958) (50, 6.636940) (55, 6.771739) (60, 6.639108) (65, 6.528958) (70, 6.797101) (75, 6.635870) (80, 6.598485) (85, 6.471591) (90, 6.480182) (95, 6.359821) (100, 6.372914) };
			
\addplot[red, mark=diamond*, mark size=1.5pt, solid, thick] coordinates {  (5, 5.824831) (10, 6.745339) (15, 5.893743) (20, 5.345899) (25, 5.144679) (30, 5.087326) (35, 5.064803) (40, 4.803505) (45, 4.741790) (50, 4.648695) (55, 4.641674) (60, 4.751316) (65, 4.702082) (70, 4.698357) (75, 4.714509) (80, 4.684733) (85, 4.602219) (90, 4.561808) (95, 4.637719) (100, 4.678009) };
			
			\end{axis}
			\end{tikzpicture}
			\caption{Misclassification rate on point clouds}
			
		\end{subfigure}\begin{subfigure}{0.5\textwidth}
			\begin{tikzpicture}
			\begin{axis}[
			xlabel={Rank $r$},
			ylabel={Runtime (s)},
			width=\linewidth,
			height = 5cm,
			every tick label/.append style={font=\footnotesize},
			xtick = {0,25,50,75,100},
			ytick = {20,40,60,80,100},
			ymin = 0,
			xmin = 0, ymax=80,
			xmajorgrids,
			ymajorgrids,
			]
\addplot[blue, mark=*, mark size=1pt, solid, thick] coordinates {  (5, 8.077686) (10, 5.791152) (15, 8.934810) (20, 10.534296) (25, 11.764867) (30, 11.075499) (35, 14.493706) (40, 16.139526) (45, 16.701329) (50, 19.005901) (55, 20.569615) (60, 22.990505) (65, 23.813586) (70, 25.401349) (75, 27.216001) (80, 26.985670) (85, 27.828160) (90, 29.154234) (95, 32.662254) (100, 34.057123) };
			
\addplot[red, mark=diamond*, mark size=1.5pt, solid, thick] coordinates {  (5, 10.350390) (10, 10.817866) (15, 14.102277) (20, 17.097813) (25, 21.307033) (30, 24.078309) (35, 26.614240) (40, 30.274297) (45, 33.858816) (50, 40.228489) (55, 44.764945) (60, 47.856402) (65, 52.394797) (70, 55.871012) (75, 58.803951) (80, 62.541126) (85, 62.674101) (90, 70.546428) (95, 73.636951) (100, 76.324251) };

			\end{axis}
			\end{tikzpicture}
			\caption{Runtime on point clouds}
			
		\end{subfigure}

		\begin{tikzpicture}
		\begin{axis}[
		width=\linewidth,
		height=2cm,
		hide axis,
		xmin=0, xmax=1, ymin=0, ymax=1,
		font=\footnotesize,
		legend columns = -1,
		legend style={
			legend cell align=left, 
			at={(0.5,0.5)}, 
			anchor=center,
			/tikz/every even column/.append style={column sep=0.5cm}},
		xtick = {5,20,50},
		ymajorgrids,
		]
		
		\addlegendimage{darkgreen, mark=square*, mark size=1pt, solid, thick}
		\addlegendentry{Mushroom}
		
		\addlegendimage{amber, mark=triangle*, mark size=1.5pt, solid, thick}
		\addlegendentry{Covertype45}
		
		\addlegendimage{magenta, mark=triangle*, mark size=1.5pt, solid, thick, mark options={rotate=180}}
		\addlegendentry{Covertype67}
		
		\end{axis}
		\end{tikzpicture}
		
		\begin{subfigure}{0.5\textwidth}
			
			\begin{tikzpicture}
			\begin{axis}[
			xlabel={Rank $r$},
			ylabel={Misclassification rate (\%)},
			width=\linewidth,
			height = 5cm,
			every tick label/.append style={font=\footnotesize},
			ymode = log,
			xtick = {0,25,50,75,100,125},
			log ticks with fixed point,
ytick = {0.4,1,4,10},
			minor ytick = {0.3,0.5,0.6,0.7,0.8,0.9,2,3,4,5,6,7,8,9},
			yminorgrids=false,
			xmajorgrids,
			ymajorgrids,
			xmin = 0
			]
			
\addplot[darkgreen, mark=square*, mark size=1pt, solid, thick] coordinates {  (5, 11.325765) (10, 10.191634) (15, 8.593411) (20, 8.515301) (25, 8.485316) (30, 8.469521) (35, 8.427813) (40, 8.445706) (45, 8.447927) (50, 8.803554) (55, 8.465202) (60, 8.607601) (65, 8.715819) (70, 8.216313) (75, 8.503702) (80, 8.428554) (85, 8.317498) (90, 8.529121) (95, 8.463228) (100, 8.734822) (105, 8.572187) (111, 8.615622) };
			
\addplot[amber, mark=triangle*, mark size=1.5pt, solid, thick] coordinates {  (5, 0.534943) (10, 0.730360) (15, 0.660311) (20, 0.648527) (25, 0.575941) (30, 0.583142) (35, 0.573895) (40, 0.605074) (45, 0.629787) (50, 0.635843) (55, 0.583961) (60, 0.623486) (65, 0.736907) (70, 0.625614) (75, 0.549836) (80, 0.604910) (85, 0.467512) (90, 0.462602) (95, 0.449673) (100, 0.433961) (103, 0.415712) };
			
\addplot[magenta, mark=triangle*, mark size=1.5pt, solid, thick, mark options={rotate=180}] coordinates {  (5, 5.019838) (10, 3.803286) (15, 4.207940) (20, 3.910769) (25, 4.114510) (30, 4.173125) (35, 4.287160) (40, 4.351216) (45, 4.382149) (50, 4.412183) (55, 4.357318) (60, 4.363288) (65, 4.315080) (70, 4.331775) (75, 4.315186) (80, 3.982909) (85, 4.203635) (90, 3.865256) (95, 3.860950) (100, 3.847558) (105, 3.786856) (110, 3.798980) (115, 3.671844) (120, 3.717252) (124, 3.671025) };

			\end{axis}
			\end{tikzpicture}
			\caption{Misclassification rate on hypergraphs}
		\end{subfigure}\begin{subfigure}{0.5\textwidth}
			
			\begin{tikzpicture}
			\begin{axis}[
			xlabel={Rank $r$},
			ylabel={Runtime (s)},
			width=\linewidth,
			height = 5cm,
			every tick label/.append style={font=\footnotesize},
			xtick = {0,25,50,75,100,125},
			ytick = {1,2,3,4,5},
			xmin = 0,
			ymin = 0, ymax = 4,
			xmajorgrids,
			ymajorgrids,
			]

\addplot[darkgreen, mark=square*, mark size=1pt, solid, thick] coordinates {  (5, 3.217430) (10, 3.094404) (15, 3.120401) (20, 3.097452) (25, 3.096356) (30, 3.064809) (35, 3.097559) (40, 3.085567) (45, 3.086989) (50, 3.080604) (55, 3.096346) (60, 3.066796) (65, 3.084912) (70, 3.079691) (75, 3.082656) (80, 3.087524) (85, 3.066329) (90, 3.071839) (95, 3.076014) (100, 3.070389) (105, 3.058658) (111, 3.147456) };
			
\addplot[amber, mark=triangle*, mark size=1.5pt, solid, thick] coordinates {  (5, 3.101961) (10, 3.070379) (15, 3.047764) (20, 3.079433) (25, 3.071698) (30, 3.061834) (35, 3.050351) (40, 3.074694) (45, 3.085560) (50, 3.067278) (55, 3.079266) (60, 3.058808) (65, 3.069720) (70, 3.042389) (75, 3.074602) (80, 3.062056) (85, 3.048136) (90, 3.080386) (95, 3.079308) (100, 3.074148) (103, 3.126265) };
			
\addplot[magenta, mark=triangle*, mark size=1.5pt, solid, thick, mark options={rotate=180}] coordinates {  (5, 3.329377) (10, 3.382393) (15, 3.343351) (20, 3.336765) (25, 3.322012) (30, 3.319963) (35, 3.391594) (40, 3.267224) (45, 3.380918) (50, 3.378337) (55, 3.390170) (60, 3.291860) (65, 3.417770) (70, 3.433058) (75, 3.412120) (80, 3.456662) (85, 3.335288) (90, 3.393945) (95, 3.437509) (100, 3.413906) (105, 3.451185) (110, 3.425149) (115, 3.540437) (120, 3.460726) (124, 3.465760) };

			\end{axis}
			\end{tikzpicture}
			\caption{Runtime on hypergraphs}
		\end{subfigure}
		
	\end{center}
	
	\caption{Misclassification rate and runtime development over different ranks $r$}
	\label{fig:rankdependency}
	
\end{figure}

Since the target rank $r$ is the only metaparameter of our method, it is important to discuss its impact. As is typical for low-rank approximations, the choice of rank is subject to a trade-off between runtime and accuracy. Figure~\ref{fig:rankdependency} depicts the development of the misclassification rate (plotted logarithmically) and runtime.

We note that the accuracy depends nontrivially on the rank. For almost all datasets, the best performance is achieved with the highest tested rank, but the dependency is not monotonous. Especially the hypergraphs show behavior where increasing the rank over a short range slightly deteriorates the accuracy. The point clouds are closer to monotony in this regard. However, the method appears to be very robust with respect to the choice of $r$, as long as it is not too small ($r < 20$).

The runtime development, on the other hand, depends on the exploited structure of the adjacency matrix. The total effort is governed by the setup cost, while the cost of layer operations does not seem to have a big impact. For point clouds, the number of computed eigenvalues influences the number of Krylov-Schur iterations, which leads to an almost linear dependency on $r$. For hypergraphs, the SVD of the normalized adjacency matrix is cheap enough to avoid any increase in runtime, leading to almost constant times.

For the best performance, we recommend choosing the maximum $r=|E|-1$ on hypergraphs without worrying about this parameter. For point clouds, we encourage choosing $r$ as large as possible while keeping a practically feasible computational cost.

\subsection{Analysis of learned weight entries}

As described in Section~\ref{sec:framework:featuremap}, our neural network learns the parameters of the filter functions \eqref{eq:filter3d} in training. The individual learned filters \eqref{eq:learnedfilter} may put varying focus on each of the three parts \eqref{eq:pgcn:zeroimpulsepart}--\eqref{eq:pgcn:highpasspart} as determined by the magnitudes of $W_{ij}^{(1,l)},W_{ij}^{(2,l)},W_{ij}^{(3,l)}$. By forming the averages of the absolute weight entries, we can quantify the importance of each filter part for the trained network. To account for the different weight matrix sizes in each layer, we use the formula
\begin{equation}
	\mu_k = \frac{1}{2} \left( \frac{1}{dh} \sum_{i=1}^d \sum_{j=1}^h |W_{ij}^{(k,1)}| + \frac{1}{hm} \sum_{i=1}^h \sum_{j=1}^m |W_{ij}^{(k,2)}| \right) \qquad (k=1,2,3).
\end{equation}
These numbers are furthermore averaged over all 100 runs.
Table~\ref{tbl:avgweights} lists these values for multiple PinvGCN instances. We clearly see that the pseudoinverse part consistently is the most important filter basis function, which supports the notion that these eigenvectors carry the most clustering information. The weights of the other two parts are smaller, but not by a large margin, which supports the intuition that the other eigenvectors are still beneficial for the classification result. At the same time, we observe that the entry ratios differ quite a bit between datasets, which implies that is indeed hard to manually choose parameters for one suitable filter function as in \eqref{eq:filter3d} a priori.

\begin{table}
	\caption{Average absolute entries in weight matrices for different filter basis functions}
	\label{tbl:avgweights}
	\begin{center}
		\begin{tabular}{@{}lcccc@{}}
			\toprule
			\multirow{2}{*}{Dataset} & \multirow{2}{*}{Rank} & Part 1 & Part 2 & Part 3 \\
			&& (zero-impulse) & (pseudoinverse) & (high-pass) \\
			\midrule
			Oakland 1, $\sigma=100$ & \multirow{2}{*}{100} & 0.085 & 0.488 & 0.193 \\
			Oakland 2, $\sigma=100$ & & 0.247 & 0.464 & 0.191 \\[0.2em]
			Mushroom & & 0.075 & 0.137 & 0.046 \\
			Covertype45 & $|E|-1$ & 0.053 & 0.112 & 0.013 \\
			Covertype67 & & 0.034 & 0.114 & 0.009 \\
			\bottomrule
		\end{tabular}
	\end{center}
\end{table}

\section{Conclusion}

We here presented Pseudoinverse GCN, a new type of Graph Convolutional Network designed for dense graphs and hypergraphs. The feature maps are based on a novel three-part filter space motivated by a low-rank approximation of the Laplacian pseudoinverse.
The method yielded strong experimental results in a setting where popular GNNs struggle. A further advantage of our method is the robustness with respect to its only parameter.
Future work might include extensions towards supervised 3D point cloud segmentation.

\bibliographystyle{spbasic}      
\bibliography{pinvgcn-refs}   

\end{document}